\documentclass[fleqn,10pt]{wlscirep}
\usepackage[utf8]{inputenc}
\usepackage[T1]{fontenc}
\usepackage{multirow}
\usepackage{dblfloatfix}
\usepackage{textcomp}
\usepackage[ruled]{algorithm2e}
\usepackage{booktabs}
\usepackage{adjustbox} 
\usepackage{xcolor}
\usepackage{amsfonts} 
\usepackage{makecell}

\usepackage[acronym]{glossaries}
\newacronym{ann}{ANN}{Artificial Neural Network}
\newacronym{cnn}{CNN}{Convolutional Neural Network}
\newacronym{rnn}{RNN}{Recurrent Neural Network}
\newacronym{snn}{SNN}{Spiking Neural Network}
\newacronym{scnn}{SCNN}{Spiking Convolutional Neural Network}
\newacronym{bptt}{BPTT}{backpropagation through time}
\newacronym{dvs}{DVS}{dynamic vision sensor}

\newcommand{\tmax}{$T_\text{max}$\xspace}
\newcommand{\bl}{\textbf}

\title{Ultra-low-power Image Classification on Neuromorphic Hardware}

\author[1,2,*]{Gregor Lenz}
\author[3]{Garrick Orchard}
\author[1]{Sadique Sheik}
\affil[1]{SynSense AG, Zurich, Switzerland}
\affil[2]{Neurobus, Toulouse, France}
\affil[3]{Intel Labs, Santa Clara, USA}
\affil[*]{mail@lenzgregor.com}

\begin{abstract}
Spiking neural networks (SNNs) promise ultra-low-power applications by exploiting temporal and spatial sparsity. The number of binary activations, called spikes, is proportional to the power consumed when executed on neuromorphic hardware. Training such SNNs using backpropagation through time for vision tasks that rely mainly on spatial features is computationally costly. Training a stateless artificial neural network (ANN) to then convert the weights to an SNN is a straightforward alternative when it comes to image recognition datasets. Most conversion methods rely on rate coding in the SNN to represent ANN activation, which uses enormous amounts of spikes and, therefore, energy to encode information. Recently, temporal conversion methods have shown promising results requiring significantly fewer spikes per neuron, but sometimes complex neuron models. We propose a temporal ANN-to-SNN conversion method, which we call Quartz, that is based on the time to first spike (TTFS).
Quartz achieves high classification accuracy and can be easily implemented on neuromorphic hardware while using the least amount of synaptic operations and memory accesses. It incurs a cost of two additional synapses per neuron compared to previous temporal conversion methods, which are readily available on neuromorphic hardware. We benchmark Quartz on MNIST, CIFAR10, and ImageNet in simulation to show the benefits of our method and follow up with an implementation on Loihi, a neuromorphic chip by Intel.
We provide evidence that temporal coding has advantages in terms of power consumption, throughput, and latency for similar classification accuracy. Our code and models are publicly available.
\end{abstract}
\begin{document}

\flushbottom
\maketitle
%
%
\thispagestyle{empty}

\section{Introduction}
\label{sec:introduction}

Deep learning models are scaling up quickly, doubling their number of parameters as frequently as every three months on average~\cite{openai2018ai}. In their pursuit of better performance, they also become ever more power-hungry.
Spiking Neural Networks (SNNs) are a class of networks that exploit high temporal and spatial activation sparsity with the goal of reducing power consumption when executed on neuromorphic hardware.

SNNs are a subclass of recurrent networks and therefore work with sequential input. For tasks such as object recognition or face detection, which rely mostly on spatial features, it is not necessary to train an SNN using backpropagation through time, which scales up training time by the number of time steps. Apart from the training cost itself, SNN training using surrogate gradients also involves certain approximations, and standard procedures in Artificial Neural Networks (ANNs) to train deeper architectures such as batch normalization are not as straightforward to apply.
A much more economical training method for SNNs when dealing with spatial tasks is to train an ANN directly and then transfer its parameters to an SNN for efficient inference. Here, the original input and activation will be distributed over multiple time steps, which are then fed to the SNN in the hope of saving costly ANN multiply-accumulate operations.

Such conversion methods have predominantly relied on rate-coding so far~\cite{diehl2015fast, rueckauer2017conversion, hu2018spiking, sengupta2019going, rueckauer2021nxtf, liu2022spikeconverter, bu2021optimal}. That means that the continuous activation value of an ANN is encoded in a number of spikes $n$ along $t$ time steps. This conversion method is straightforward to implement, robust to firing time errors and works on neuromorphic hardware~\cite{massa2020efficient, rueckauer2021nxtf}. On the flip side it requires an enormous amount of spikes to encode information while still running into issues of not being able to propagate the signal to deeper layers~\cite{rueckauer2016theory, rueckauer2017conversion}. Since the number of spikes is directly linked to energy consumption on a neuromorphic chip, this approach has difficulties outperforming the original ANNs in terms of efficiency~\cite{davidson2021comparison}.

In contrast, conversion frameworks based on temporal coding have made significant advances as of late, at much fewer spikes in comparison to rate coded SNNs~\cite{zhang2019tdsnn, han2020deep, chu2020you, park2020t2fsnn}. 
They make use of the exact timing of spikes when encoding the original activation, which drastically reduces the number of spikes needed. We obtain a number of advantages over rate coding: 

\begin{itemize}
    \item Each layer's activities are decoupled from each other.  The input signal does not need to be presented for an extended time to propagate to later layers and neurons do not need to undergo a strong transient response resulting in uneven firing rates across time. Generally, temporal methods use much fewer spikes.
    \item Naturally suitable for max-pooling, as the maximum output is the first neuron that fires. A refractory period or self-inhibitory connections suppress subsequent outputs. In rate coding, a maximum value has to be determined over a time window, which is more difficult. 
    \item Biases are encoded using the timing of a single spike rather than a constant spike rate, which results in much fewer spikes.
    \item No \emph{soft reset} of membrane potentials needed to achieve good results. A soft reset is when the spiking threshold is subtracted from the membrane potential after a spike, rather than setting the membrane potential to zero. On digital systems that might receive multiple large inputs in one time step, this helps to preserve some of the information and bias the neuron in the next time step. In temporal coding methods, neurons generally fire only once, hence the reset mechanism is of no importance.
\end{itemize}

Time To First Spike (TTFS) encoding is the simplest such method which uses a single spike per neuron, potentially leading to great energy efficiency~\cite{rueckauer2018conversion}. This method translates the floating point input of an ANN into an input current of an SNN over time. If that current is large, the neuron will spike earlier and vice versa, hence time to first spike. In practice, many input currents will be integrated over time and spiking neurons struggle with early firing if the threshold is not adjusted dynamically, which lowers SNN accuracy.
Recently, a promising alternative temporal approach using sequential binary coding has been shown to achieve good performance~\cite{stockl2021optimized, rueckauer2021temporal}. This approach resides somewhat between pure rate and single spike coding in that a single floating point activation of the ANN is represented as series of binary spikes in the SNN where one time step represents a specific bit. This method is able to approximate activation functions to a very high degree.

What many temporal conversion methods have in common is that they rely on complex neuron and/or spike models, which are either not supported or at least very costly to implement on neuromorphic hardware. Most TTFS methods use dynamic neuron membrane thresholds to prevent early firing~\cite{rueckauer2018conversion, park2020t2fsnn}, and methods that use binary coding potentially face memory issues and the need for multiplication operations not available on neuromorphic cores~\cite{stockl2021optimized, rueckauer2021temporal}. Furthermore, no network structure has been proposed so far to encode SNN outputs into a binary spike train.

We propose a new temporal conversion method which we name Quartz, where the precise timing of a single spike encodes information efficiently. Rather than requiring complex neuron models or computation operations, our method works with simple neuron and synapse models and relies instead on some additional connections in the network to improve performance. 
Quartz uses TTFS encoding of activations, non-leaky integrate-and-fire neurons and synaptic currents which are readily available on neuromorphic hardware. For every neuron in the network we add two additional synapses. One synapse helps to prevent early firing and reduces quantization error, while the second synapse forces a neuron to spike if it hasn't done so at a specific time point, which is our temporal equivalent of a rectifying operation.

The main difference of Quartz in comparison to other temporal methods is that it relies on a balancing second spike per neuron at pre-defined time steps, which act as a counterweight and provide predictable readout currents. In comparison to inverted TTFS~\cite{rueckauer2018conversion}, this relieves us of the need to tweak spiking thresholds for each unit individually and also drastically reduces latency for low activations (spike that encodes zero arrives at a pre-determined step rather than at the last global time step). That also means that activations are guaranteed to arrive in later layers because every neuron will spike exactly once, whereas later layers in rate-coded networks might 'starve' if not enough neurons are activated.
It also allows us to treat the activation of each ANN layer separately within a pre-defined number of time steps, which results in further energy optimizations of the hardware.

Conversion methods based on similar temporal coding schemes~\cite{han2020deep, stockl2019recognizing, stockl2020classifying} have achieved very good accuracy, although they have yet to show that they can be implemented on neuromorphic hardware. 
Our goal is to close the accuracy gap between ANNs and converted SNNs while keeping spike count and therefore energy consumption to a minimum. 
So far, mostly networks obtained through rate-coded conversion schemes have been benchmarked on neuromorphic hardware. Using Quartz, we show simulation results for deep networks trained on MNIST, CIFAR10 as well as ImageNet  and follow up with a benchmark on Intel's neuromorphic hardware chip called Loihi for the first two datasets. 

\begin{figure*}[!bh]
    \centering
    \includegraphics[width=\linewidth]{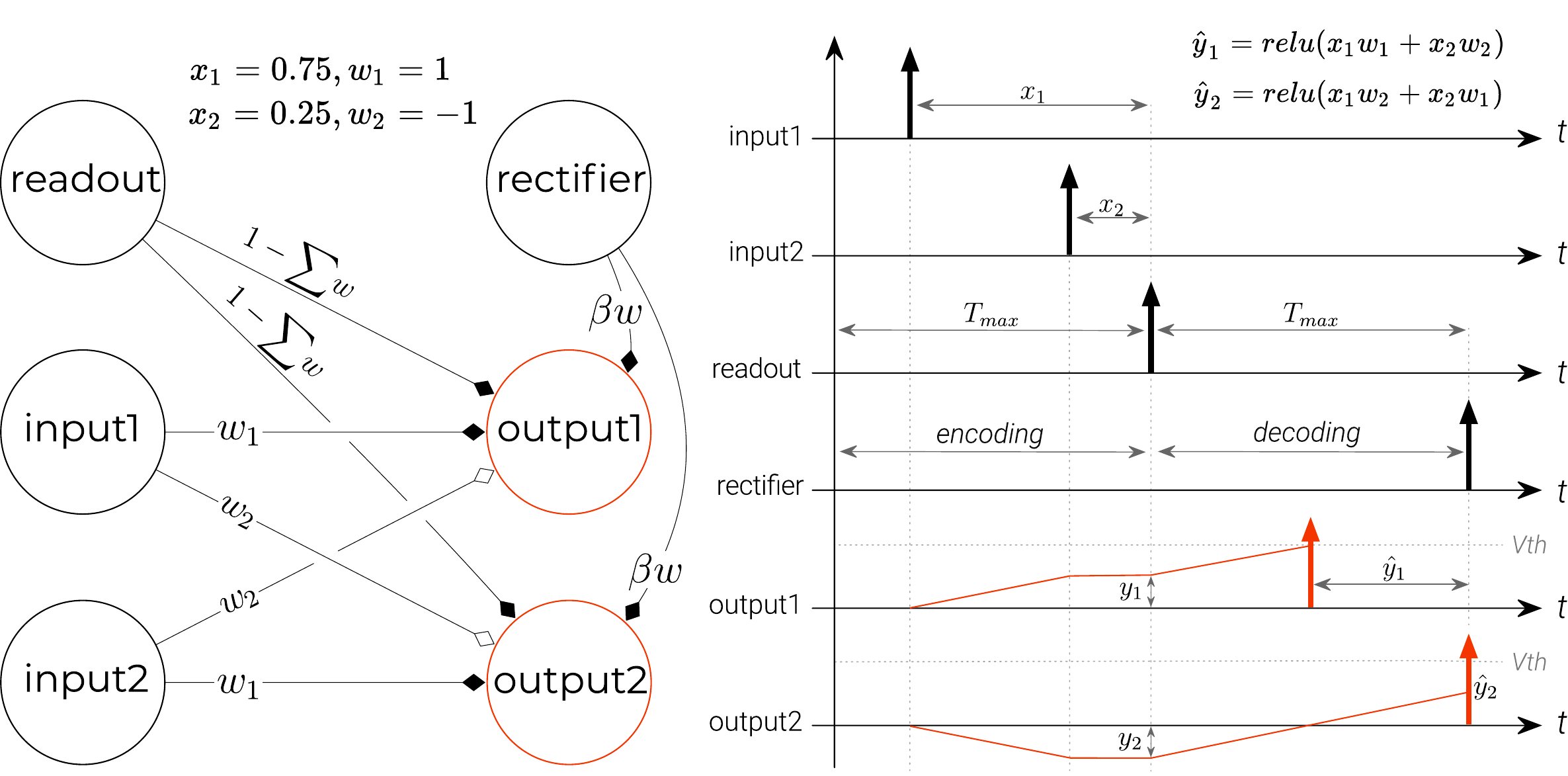}
    \caption{Quartz conversion scheme shown for 2 ANN units \emph{output1}, \emph{output2} that have been converted to spiking neurons.
    \textbf{Left:} Connection architecture. 
    The normalized ANN weights $w_{\{1,2\}}$ can be used as is to connect convolutional or fully connected layers. Inputs $x_{\{1,2\}}$ are encoded using latency coding in Eq.~\ref{eq:encoding}.
    The \emph{rectifier} injects a large current with $\beta \gg \sum w$ to force a neuron to spike if it hasn't yet at the last time step of a layer. The neuron \emph{output1} computes $\hat{y}_1 = \text{max}(0, w_1 x_1 + w_2 x_2)$, whereas \emph{output2} computes $\hat{y}_2 = \text{max}(0, w_2 x_1 + w_1 x_2)$. 
    \textbf{Right:} Chronogram of the same network, with example inputs $x_1 = 0.75, x_2 = 0.25$ and weights $w_1=1, w_2=-1$.
    As soon as input spikes arrive at the output neurons, $i(t)$ increases according to the input weights (not shown) and $u(t)$ (shown in red) starts to ramp up. After \tmax time steps, the encoding phase is completed and $u(t)$ now represents the value that neuron is supposed to output. For the next \tmax time steps, we decode that membrane potential into a spike time. The \emph{readout} neuron ensures that all input currents for a neuron are balanced by injecting a current that is the negative sum of all inputs plus a constant. 
    Whereas \emph{output1} outputs the expected $1\times0.75 - 1\times0.25=0.5$, \emph{output2} with $1\times0.25 - 1\times0.75=-0.5$  is forced to spike early by injecting a high current at time step $2 T_\text{max}$. This spike coincides with the \emph{readout} of the next layer (not shown here), where their effects will cancel out because the output is $0$. For this diagram $T_\text{max}$ is assumed to be  large such that transmission delays are negligible.}
    \label{fig:ann-snn-conversion}
\end{figure*}

\section{Time to First Spike Conversion}
To run inference on an SNN that has been converted from an ANN, we replace ReLU activations in the ANN with our spiking layers. The normalized input data is converted to a temporal equivalent through latency coding, where we assign a spike timestamp $s$ to every normalised pixel value $p$. Brighter pixels fire earlier:
\begin{equation}
    s = \lfloor T_\text{max} (1-p) \rfloor
    \label{eq:encoding}
\end{equation}
where $T_\text{max} \in \mathcal{N}$ is the amount of time steps per layer to encode our input. The parameter is a trade-off between accuracy and latency. More time steps will mean less quantization error from the original ANN activation, but the solution will also take longer to compute.

Parameter layers will weight the input spikes and feed synaptic currents at times proportional to the value to our non-leaky integrate and fire neurons. This first part is the encoding phase, where all inputs are integrated as $i(t)$. 
After $T_\text{max}$ time steps, we set $i(t)$ to a constant readout current for the decoding. Decoding layer $l$ is encoding layer $l+1$ at the same time. 

\begin{equation}
    i(t) = 
    \begin{cases}
      b \text{H}(t- s_b) + \sum_{j} w_j \text{H}(t- s_j), & \text{if}\ t \leq T_\text{max} \\
      1, & \text{if}\ t > T_\text{max}
    \end{cases}
    \label{eq:syn-current}
\end{equation}
where $w$ are the weights, $b$ is the bias and H the Heaviside step function. 
Synaptic current is integrated onto the membrane potential $u$ at every time step:
\begin{equation}
        u(t) = \sum_{t=0} i(t)
        \label{eq:membrane-equation}
\end{equation}
with spike firing condition
\begin{equation}
    s_i = t, u_i(t) = 0 \ \text{if} \ u_i(t) \geq \theta
    \label{eq:theta}
\end{equation}
and potentially a rectifying condition
\begin{equation}
    s_i = 2T_\text{max}, u_i(2T_\text{max}) = 0 \ \text{if} \ u_i(2T_\text{max}) < \theta
    \label{eq:rectifying-jump-condition}
\end{equation}
for neuron $i$ where $\theta$ is the neuron's firing threshold. 
The second condition is optional depending on if the original ANN used a rectified activation (ReLU) or not. In hidden layers this is typically the case but can be omitted in the last or bottleneck layers to allow for negative activations. If applied, a refractory period of $T_\text{max}$ should be set for the neuron to prevent it from firing a second time. 
A spike $s > 2T_\text{max}$ will be counted as negative value. 
This leads to $u(T_\text{max}) = \sum_j w_j p_j + b$ if the firing threshold has not been exceeded prematurely.
To find out when the neuron spiked we solve for $u(t) = \theta$:

\begin{equation}
    \theta = \sum_{t=0}^{T_\text{max}} i(t) + t - T_\text{max}
\end{equation}
\begin{equation}
    t = \theta + T_\text{max} - u(T_\text{max}) 
\end{equation}

which can be simplified to
\begin{equation}
    t = 2T_\text{max} - u(T_\text{max})
\end{equation}
for $\theta = T_\text{max}$.
A network scheme and chronogram of our model can be seen in Figure~\ref{fig:ann-snn-conversion}.
For each layer in the original ANN, we receive an input and calculate an activation. In the converted SNN, those values are treated over time in \emph{encoding} and \emph{decoding} phases that each have a pre-determined number of \tmax time steps. This mechanism also allows for higher throughput by feeding a new input every $2T_\text{max}$ time steps much like in other recent temporal conversion works~\cite{stockl2021optimized, rueckauer2021temporal}. 


\begin{figure}[ht]
    \centering
    \includegraphics[width=0.8 \columnwidth]{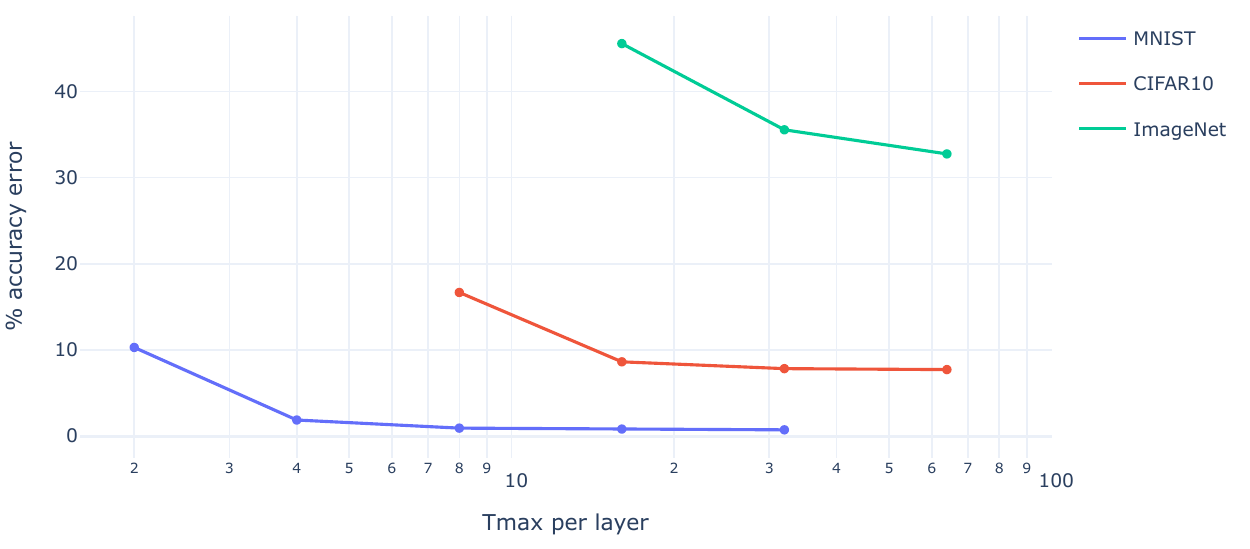}
    \caption{Classification accuracy error of SNN converted using Quartz as a function of $T_\text{max}$ time steps per layer. Choosing a larger number of time steps will reduce quantization error but increase latency. x axis is plotted logarithmically from \tmax of $1$ to $100$.}
    \label{fig:simulation-t_max}
\end{figure}

\section{Results in Simulation}
We implement Quartz to test SNNs that have been converted from pre-trained ANNs with full (32 bit) weight resolution. The results for image classification datasets MNIST~\cite{lecun1998mnist}, CIFAR10~\cite{krizhevsky2009learning} and ImageNet~\cite{krizhevsky2017imagenet} are shown in Table~\ref{tab:results-simulation}.
We can trade-off numerical accuracy in the network against number of time steps to compute using $T_\text{max}$. Figure~\ref{fig:simulation-t_max} shows how choosing a number of time steps per layer influences quantization error, where lower \tmax will drive up the error and impact classification accuracy. The whole network uses $T_\text{max}l$ time steps to classify one input, where $l$ is the number of layers. 
All the neurons in Quartz only fire once per sample.
The conversion framework as well as pre-trained models are publicly available (see data availability statement).

\begin{table}[!htb]
    \centering
    
    \caption{Classification performance for different conversion methods on MNIST, CIFAR10 and ImageNet.}
    \begin{tabular}{cccccccccc}
    \toprule
    \bl{Dataset} & \bl{Architecture} & \bl{Params} & \bl{Method} & \bl{\makecell{ANN \\ error [\%]}} & \bl{\makecell{SNN \\ error [\%]}} & \bl{Neurons} & \bl{Spikes} & \bl{\makecell{Time\\steps}} & \bl{Ops} \\ 

    \midrule
    \multirow{6}{*}{MNIST} 
    &    -     & 760k & Rate~\cite{zambrano2017efficient}    & 0.41 & 0.49 &  29k & 120k &  - & 160M \\
    & LeNet-5 & 213k & Rate~\cite{rueckauer2021temporal}    & 0.72 & 1.16 &   8k &   4k &  - & 1.8M \\
    & LeNet-5 & 213k &Latency~\cite{rueckauer2018conversion}& 1.04 & 1.43 & 7.6k &   1k &  - & 230k \\
    &    -     & 213k & Latency~\cite{zhang2019tdsnn}        & 0.84 & 0.92 &   8k &   -  &  - & - \\
    & LeNet-5 & 213k & Pattern~\cite{rueckauer2021temporal} & 0.72 & 1.26 &   8k &   2k &  - & 865k \\
    &\bl{LeNet-5}&\bl{26k}&\bl{Quartz}      &\bl{0.75}&\bl{0.80}&\bl{4.4k}&\bl{4.4k}&\bl{50}&\bl{177k} \\

    \midrule
    \multirow{10}{*}{CIFAR10} 
    &      -       &   9M & Rate \cite{zambrano2017efficient}   & 10.34 & 10.33 & 182k &   4M &  -   &4700M \\
    & MobileNet-V1&   3M & Rate~\cite{rueckauer2021temporal}   &  8.82 &  9.40 & 413k &   5M &  -  & 576M \\
    &      -       & 118k & Pattern \cite{kim2018deep}          & 10.90 & 10.90 & 118k & 100M &   -  &   -  \\
    &      -       &   -   & Burst \cite{park2019fast}           &  8.59 &  8.59 & 281k &   7M &   -  &   -  \\
    &       -      &   -   & TSC~\cite{han2020deep}              &  8.53 &  8.58 &    - &    - & -   &   -   \\
    & ResNet-20   &   -   & Pattern~\cite{stockl2021optimized}  &  8.42 &  8.55 &  -    &   -   & 200 &    -  \\
    & ResNet-50   & 0.8M & Pattern~\cite{stockl2021optimized}  &  7.01 &  7.58 & 475k & 700k & 500 &   -  \\
    & MobileNet-V1&   3M & Pattern~\cite{rueckauer2021temporal}&  8.82 &  8.82 & 413k &   2M &  -  &  39M \\
    &\bl{VGG-11} &\bl{28M}&\bl{Quartz}    &\bl{7.61} &\bl{7.77} &\bl{160k} &\bl{160k} &\bl{258} &\bl{12M} \\
    &\bl{VGG-11} &\bl{28M}&\bl{Quartz}    &\bl{7.61} &\bl{7.62} &\bl{160k} &\bl{160k} &\bl{514} &\bl{22M} \\
    &\bl{MobileNet-V1}&\bl{2M}&\bl{Quartz} &\bl{11.41}&\bl{13.40}   &\bl{313k} &\bl{313k} &\bl{1153}&\bl{40M}\\

    \midrule
    \multirow{6}{*}{ImageNet} 
    &    -         &   -  & Rate \cite{zambrano2017efficient}  & 37.02 & 37.11 &   2M & 110M &  -  & -\\
    & MobileNet-V1 &   4M & Rate~\cite{rueckauer2021temporal}  & 30.41 & 36.01 &   5M &  75M &  -  & 24G \\
    & VGG-16       & 138M & Latency \cite{zhang2019tdsnn}      & 30.82 & 29.13 &  15M &    - &  -  & 17G \\
    & ResNet-50    & 0.8M & Pattern~\cite{stockl2021optimized} & 24.78 & 24.90 &  10M &  14M & 500 & -\\
    & MobileNet-V1 &   4M &Pattern~\cite{rueckauer2021temporal}& 30.41 & 31.13 &   5M &  15M &  -  & 3.6G \\
    &\bl{VGG-11}  &\bl{28M} &\bl{Quartz}  &\bl{31.89} &\bl{32.66} &\bl{7.4M} &\bl{7.4M} &\bl{514} &\bl{1.2G}\\

    \bottomrule
    \end{tabular}
    \label{tab:results-simulation}
\end{table}

\subsection{MNIST} We train a smaller version of LeNet-5 and  achieve the lowest absolute classification error (0.8\%), the lowest drop in accuracy during conversion ($0.05$\%) and the lowest number of operations per inference (177k) compared to any previously reported  method. Whereas the full inference takes 64 time steps, the first classification spike is received after 50 time steps on average. 
The number of synaptic operations, defined as spikes times the neuron fanout, is 38k. 

\subsection{CIFAR10} We observe a drop of as little as $0.01$\% when converting a pre-trained VGG-11 network with $T_\text{max} = 64$, at 514 time steps per inference. Decreasing \tmax to 32 reduces latency to 258 time steps on average, while conversion error increases to 0.16\%. The number of synaptic operations in the second case is 12M. We also benchmark a MobileNet-V1 architecture, which has more than twice the amount of layers as VGG-11. This increases the amounts of time steps considerably, as well as quantization error and number of operations. Using Quartz, we achieve a reduction in number of spikes by an order of magnitude using the VGG architecture compared to previously reported methods, and the total number of operations is the lowest reported. The first layers in our converted networks are processed as floating point values, because our activation normalization scheme doesn't take into account the standardization pre-processing usually employed in ANN training, where inputs range from large negative to positive numbers.

\subsection{ImageNet} As for CIFAR10, we use a pre-trained VGG-11 network and process the first layer as floating points to mitigate input standardization. We observe a drop of 0.87\% accuracy between the ANN and the converted SNN for a latency of 514 time steps and 1.2G operations. For an equivalent number of neurons in the network, our method uses half the amount of spikes compared to the best reported method in simulation. In comparison to rate-coded models, we reduce network computation cost by a factor of 20 while at the same time achieving lower classification error. 

\begin{figure}[htb]
    \centering
    \includegraphics[width=\columnwidth]{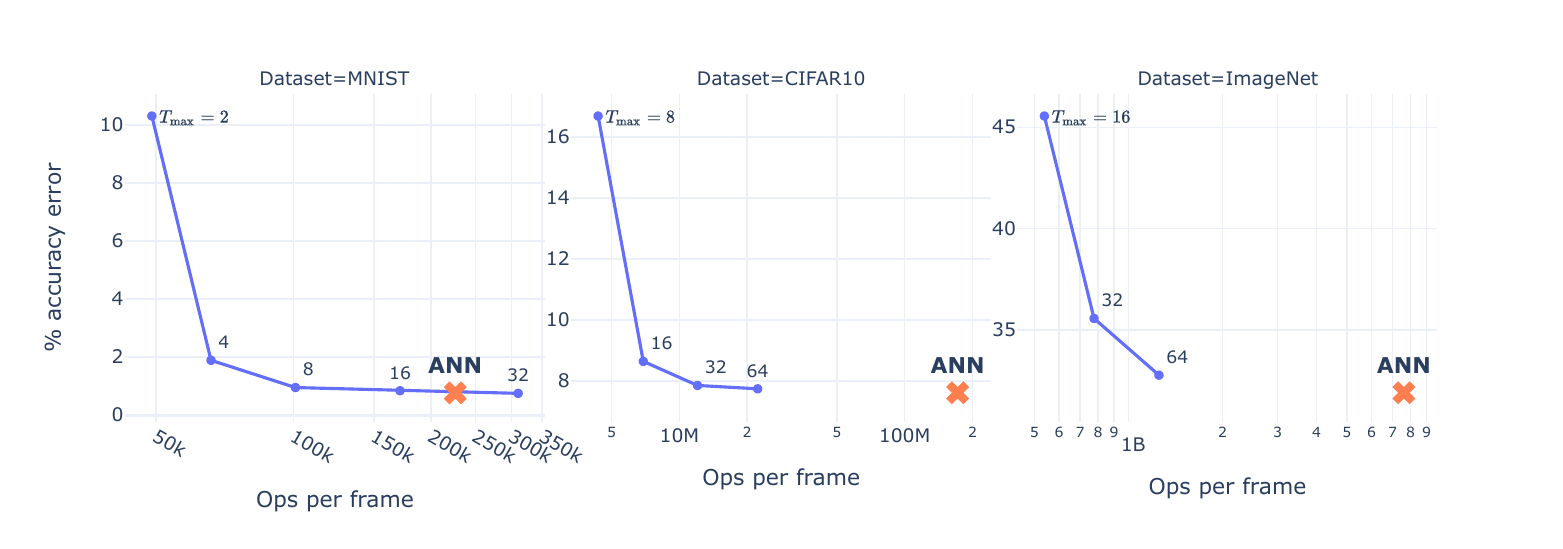}
    \caption{Classification accuracy error in percent over the number of operations per image for three different datasets. The floating point operations in the pre-trained ANNs are counted using Meta's fvcore tool. Operations in SNNs are counted according to Equation~\ref{eq:ops-count} and are additions only. By exploiting sparsity in the activation, we can drastically reduce the number of overall operations needed, and observe a error/operation trade-off depending on the amount of time steps per layer chosen ($T_\text{max}$). }
    \label{fig:n_ops}
\end{figure}

\begin{figure}[htb]
    \centering
    \includegraphics[width=\columnwidth]{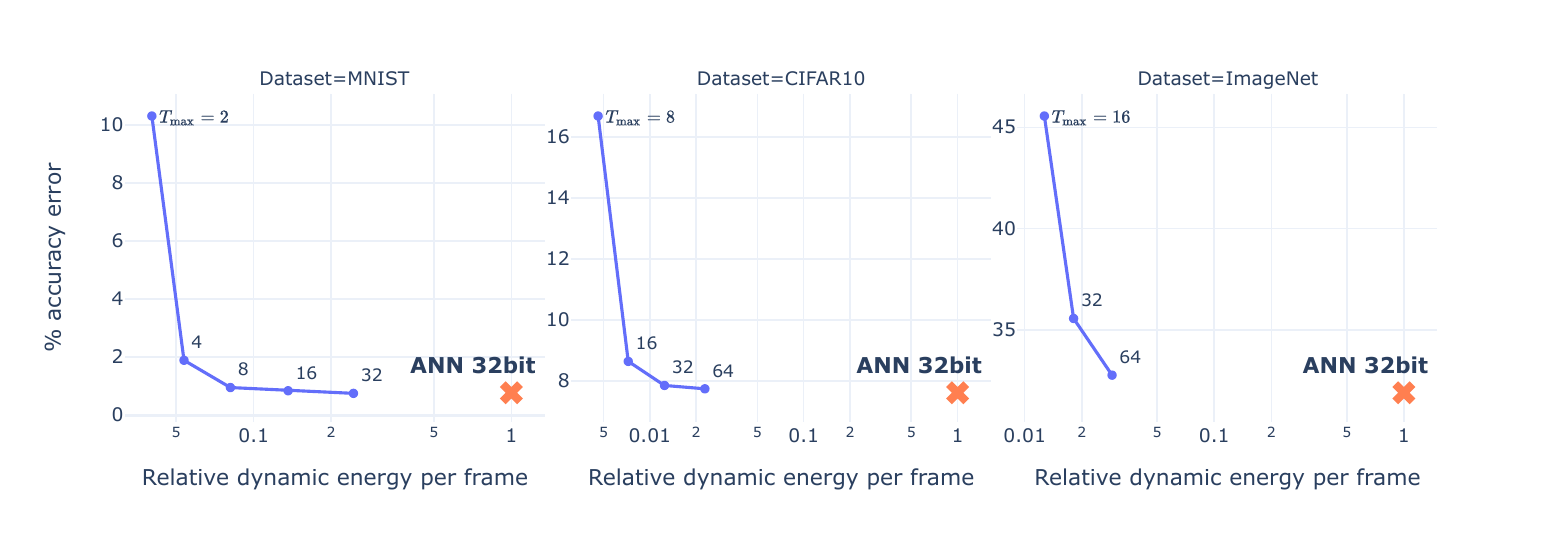}
    \caption{Weighing the number of addition or MAC operations per frame by their respective energy cost, we observe a significant reduction in dynamic energy for Quartz SNNs. Much like in Figure~\ref{fig:n_ops}, we observe a trade-off between the classification accuracy error and dynamic energy, chosen through the number of time steps per layer. }
    \label{fig:energy_comp}
\end{figure}

\subsection{Counting the Number of Operations and Estimating Power Consumption}
\label{sec:operations}
Tab.~\ref{tab:results-simulation} includes the number of operations for different networks. For Quartz networks, this number is the sum of all addition operations:
\begin{equation}
    \Omega = \sigma + \nu 2 T_\text{max}
    \label{eq:ops-count}
\end{equation}
where $\sigma$ is the number of synaptic operations (which do not include neuron updates) and accounts for adding input weights to $i(t)$. $2\nu T_\text{max}$ is an upper bound to update $u(t)$ for every neuron $\nu$.
Throughout the benchmarks, we reduce the number of operations by a factor of 3-4 compared to the best previously reported result which is binary pattern coding, and by a factor of about 20 in comparison to rate-coded methods that have been implemented on neuromorphic hardware.
The number of operations scales linearly with $T_\text{max}$.

In Figure~\ref{fig:n_ops} we compare the number of operations of Quartz networks with their original ANNs. To count the number of floating point operations in the ANN, we use Meta Research's \textit{fvcore} package~\cite{fvcore}. By distributing the sparse activation in time, we show that our SNNs use around an order of magnitude fewer operations than the original ANN. 

The number of operations does not directly compare to the dynamic power used, as one operation can be multiply-accumulate (MAC), simple addition, a bitshift or yet others. In Quartz we use addition operations only. For two operands of bitwidth b1 and b2, the cost of addition is $\text{max}(b1, b2) + \frac{b1 - b2}{2}$ and a MAC consumes $b1 b2$~\cite{rueckauer2021temporal}.  We weigh the energy required for ANNs and our converted SNNs based on the number of operations required. Figure~\ref{fig:energy_comp} shows the relative dynamic energy between these two networks, resulting in 1-2 orders of magnitude in reduction of dynamic power consumption compared to the ANN.

\section{Results on Loihi}
To measure the energy efficiency of our method on hardware, we implement Quartz on Loihi and make it publicly available (see data availability statement). Loihi is Intel's neuromorphic research processor that implements SNNs in a fully digital architecture\cite{davies2018loihi}. Its first version features 128k artificial neurons and 128M synapses across 128 neuromorphic cores per chip, and chips can be tiled to further scale up the total available resources. 
Continuous functions of membrane potentials are approximated in discrete time steps on Loihi with all neurons synchronized to a common time step throughout the entire system. 
When a neuron enters a firing state, it generates spike messages that get routed by a network on chip to target cores where the spikes fan out to synaptic connections. 
It uses 8 bit weight resolution and supports axonal delays, which schedule all outgoing spikes to arrive at a future time step and thus determine a maximum transmission delay between two neurons. 

To realize the conditions in Equations~\ref{eq:syn-current} and \ref{eq:rectifying-jump-condition}, we make use of two additional synapses per neuron. A plot of a converted network is shown in Figure~\ref{fig:ann-snn-conversion}. The pre-synaptic input from \emph{readout} neutralizes all input currents and starts the decoding from membrane potential back to a timed spike. Since all pre-synaptic neurons have fired at $T_\text{max}$ and input weights are known a priori, we can balance the input current exactly at time $T_\text{max}$. This idea is loosely based on spike time interval encoding, with the addition that the second spikes of all input pairs are aligned and can therefore be summed up in a single counter spike with weight $1 - b - \sum_{j} w_j$ at time $T_\text{max}$.

The second synapse from \emph{rectifier} will inject a large current into post-synaptic neurons that need to be rectified at time step $2T_\text{max}$.
To prevent neurons from firing a second time, we can either make use of a large enough refractory period or self-inhibitory connections. 

We show that we can successfully deploy converted models on constrained neuromorphic hardware and provide evidence that temporal encoding has several advantages in comparison to the dominant rate-coding conversion scheme. 

\begin{table}[!htb]
    \centering
    
    \caption{Comparison to other converted SNNs on neuromorphic hardware for MNIST and CIFAR10 classification. }
    \begin{tabular}{cccccccc}
    \toprule
    \textbf{Dataset} & \textbf{Method} & \textbf{\makecell{ANN \\ error [\%]}} & \textbf{\makecell{SNN \\ error [\%]}} & \textbf{spikes} & \textbf{time steps} & \textbf{neurons} & \textbf{\makecell{spiking \\ hardware}} \\ 
    \midrule
    
    \multirow{8}{*}{MNIST}
    & Rate~\cite{yin2017algorithm}   &   -  & 1.3  &   -   &  -  & 1.3k & Simul. 28\,nm \\
    & TTFS~\cite{mostafa2017fast}    &   -  & 3.02 &  135  & 167 & 1.3k & FPGA \\
    & Rate~\cite{zheng2018low}       &   -  & 10.0 &   -   &   - & 0.3k & Simul. 65\,nm \\
    & Rate~\cite{chen20184096}       &   -  & 1.4  & ~130k &   - &   2k & FinFET 10\,nm \\
    & TTFS~\cite{oh2020hardware}     &   -  & 3.1  &  162  & 256 &   1k & Simul. 0.35\,\textmu m \\
    & Rate~\cite{massa2020efficient} &   -  & 1.3  &   -   &   - &   8k & Loihi \\
    & Rate~\cite{rueckauer2021nxtf}  & 0.74 & 0.79 &   -   & 100 &   4k & Loihi \\
    & \textbf{Quartz}                & 0.75 & 0.77 &  5.4k &  65 & 5.4k & Loihi \\
    
    \midrule
    \multirow{3}{*}{CIFAR10}
    & Rate~\cite{massa2020efficient} &     - & 22.9  &  -  &  -   &  82k & Loihi \\
    & Rate~\cite{rueckauer2021nxtf}  &  8.07 &  8.52 &  -  &  400 & 413k & Loihi \\
    & \textbf{Quartz}                & 24.04 & 25.14 & 48k & 1211 &  48k & Loihi \\
    \bottomrule
    \end{tabular}
    \label{tab:results-comparison-snns}
\end{table}

\subsection{Classification Accuracy}
In Table~\ref{tab:results-comparison-snns} we list results for MNIST and CIFAR10 classification that have been benchmarked on spiking hardware.
For MNIST our trained ANN reaches a classification accuracy error of 0.73\% and for the converted SNN we observe a sweet spot of 0.77\% classification accuracy error with a delay of 4.91\,ms per inference when using $T_\text{max} = 2^4$. For this parameter setting, the first spike in the last layer is received after 65 time steps on average, whereas inference for one sample overall takes 91 time steps.
For CIFAR10 we compare to two other works that use rate coding on Loihi. While our ANN achieves 24.04\% accuracy error, we observe a 1.1\% drop in accuracy after conversion. Inference takes 365\,ms when using $T_\text{max} = 2^7$. The first spike is received on average after 1211 steps and the whole execution takes 1301 time steps.
The additional amount of time steps per inference, when compared to simulation for similar \tmax, is due to transmission delays on Loihi, which are not present in simulation.



\begin{table}[ht]
    \caption{Breakdown of static and dynamic power consumption and latency on Loihi for MNIST and CIFAR10 tasks. Additional split for neuromorphic cores and Lakemont x86 CPUs. Accuracies as reported in Table~\ref{tab:results-comparison-snns}. For CIFAR10, we also report static power consumption based on an improved placement of neurons across fewer cores that was not possible using the SDK at time of experiments. }
    \centering
    \begin{tabular}{llllllll}
    \toprule
    && \textbf{MNIST} &&&& \textbf{CIFAR10} &\\
    \textbf{Power consumption} [mW]  & x86     & Neuron   & Total & & x86     & Neuron   & Total \\ 
    \cmidrule{2-4}
    \cmidrule{6-8}
    \multicolumn{1}{r}{Static}  & 0.14    & 5.09    & 5.23  && 3    & 1271/ 63.55\textsuperscript{1} & 1274 \\
    \multicolumn{1}{r}{Dynamic} & 22.66   & 9.29    & 31.95 && 258   & 11    & 269 \\
    \multicolumn{1}{r}{Total}   & 22.79   & 14.38   & 37.18 && 261   & 1282   & 1543 \\ 
    \cmidrule{2-4}
    \cmidrule{6-8}
    \textbf{Latency} [ms]     & \multicolumn{3}{c}{4.91}  & & \multicolumn{3}{c}{365} \\
    \textbf{Time steps per inference} & \multicolumn{3}{c}{91} && \multicolumn{3}{c}{1301} \\
    \textbf{Energy per inference} [\textmu J] & \multicolumn{3}{c}{182.46} && \multicolumn{3}{c}{564} \\
    \textbf{Avg. time steps to $1^\text{st}$ spike}  & \multicolumn{3}{c}{65} &&  \multicolumn{3}{c}{1211} \\
    \textbf{EDP} [\textmu Js]                 & \multicolumn{3}{c}{0.9} && \multicolumn{3}{c}{206047/10302\textsuperscript{1}} \\
    \midrule
    \textbf{EDP Rueckauer et al.~\cite{rueckauer2021nxtf}} [\textmu Js] & \multicolumn{3}{c}{4.38} && \multicolumn{3}{c}{34926} \\
    \textbf{EDP ANN on GPU~\cite{rueckauer2021nxtf}} [\textmu Js] & \multicolumn{3}{c}{222} && \multicolumn{3}{c}{18924} \\
    \bottomrule
    \textsuperscript{1} Optimized neuron placement.
    \end{tabular}
    \label{tab:power-measurements}
\end{table}

\begin{figure}
    \centering
    \includegraphics[width=0.4\linewidth]{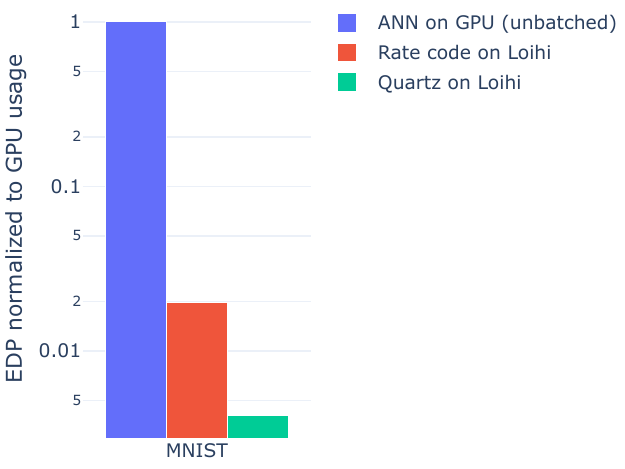}
    \includegraphics[width=0.4\linewidth]{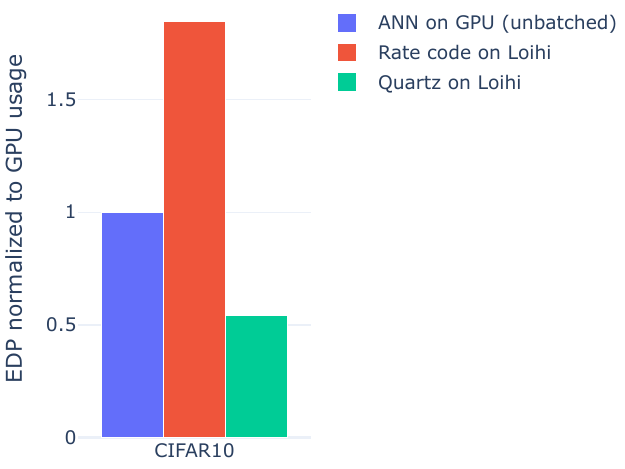}
    \caption{Energy-delay product (EDP) normalized to GPU usage, for exact numbers see Table~\ref{tab:power-measurements}.  }
    \label{fig:edp-comp}
\end{figure}

\subsection{Power Measurements}
We break down power consumption for workloads into dynamic and static components for neuromorphic cores, which store neuron states and emit spikes, and the x86 Lakemont CPUs which are responsible for executing user scripts. Static power consumption is depends to a great extent on the number of active components and the  manufacturing process. As such we focus on the consumption of dynamic energy that is consumed by switching transistors to update neuron states and route spikes and is therefore directly relevant for the workload at hand. 
Because any model also needs a certain time to compute, a preferred metric when benchmarking neuromorphic algorithm performance on hardware is the product of energy consumption and latency, the Energy Delay Product (EDP)~\cite{davies2019benchmarks}.

Table~\ref{tab:power-measurements} lists power measurement results for MNIST and CIFAR10 tasks on Loihi. All measurements were obtained using NxSDK 1.0.0 with an Intel Xeon CPU E5-2650 CPU (2.00 GHz, 64 GB RAM) as the host running Ubuntu version 20.04.2 and the Loihi Nahuku 32 board ncl-ext-ghrd-01.

For NMNIST our converted SNN model uses $T_\text{max}=2^4$ time steps per layer, occupies 11 of the 128 neuromorphic cores on one chip and consumes $182.46$\,\textmu J of energy per inference for both neuromorphic cores and Lakemont CPUs, of which $156.9$\,\textmu J is dynamic energy. We show that our EDP at  $0.9$\,\textmu Js is almost five times better compared to that of a rate-coded model at $4.38$, even without low-level connection optimizations done in Rueckauer et al.~\cite{rueckauer2021nxtf}. For comparison, inference for a single MNIST image on a GPU achieves an EDP of $222$\,\textmu Js for an error of 0.73\,\%~\cite{rueckauer2021nxtf}. 

For CIFAR10 we choose $T_\text{max} = 2^7$. Since the first version of Loihi does not provide support for weight sharing without low-level engineering effort, our SNN ends up being distributed across 1753 neuromorphic cores and 14 chips to make room for the many convolutional connections. On average we were able to use only 3\% of neurons on each core. Although dynamic energy for neuromorphic cores increases only slightly from 9.29\,mW for MNIST to 11\,mW in CIFAR10, the amount of cores that have to be powered on increases the static power consumption drastically from 5.23\,mW for MNIST to 1274\,mW for CIFAR10. 
Unfortunately Rueckauer et al.~\cite{rueckauer2021nxtf} only provide power benchmarks for static and dynamic power combined without breaking down the individual components, which makes a comparison along dynamic power impossible. We therefore estimate static power consumption of our method based on an improved placement of neurons across fewer cores as is done in Rueckauer et al.~\cite{rueckauer2021nxtf}, assuming a conservative core utilization of 60\%. The corrected static together with the original dynamic power consumption result in an EDP of 10.3\,mJs, which is 3 times lower than the reported rate-converted network and about 2 times lower than a GPU that uses 18.9\,mJs per inference~\cite{rueckauer2021nxtf}.

Overall we achieve a 2- to 5-fold improvement in EDP over rate-coded converted networks and also beat GPUs when used in single batch mode. Further improvements are possible when using the second generation of Loihi's chip as it reduces static power consumption even more and has improved support for convolutional connectivity.

\section{Discussion}
\label{sec:discussion}
Quartz is a simple method that provides temporal encoding for SNNs converted from ANNs. Conversion methods are especially suitable when spatial input features are dominant in the input data and when power consumption is key during inference for deeper networks. 
The dominant rate-coding schemes that have been implemented on neuromorphic hardware often mean employing a large number of neurons and spikes, which can scale unfavorably for bigger networks. 

We show in PyTorch simulations that we reduce the amount of operations by a factor of 3-4  compared to the best previously reported results for conversion methods, all while keeping the drop in classification accuracy to a minimum. The amount of operations relates directly to power consumption, which results in ultra-low-power image classification. 
On Loihi, we show that Quartz clearly outperforms rate-coded methods when core utilization is high enough to keep static power consumption at bay.
As recommended in Davies et al.\cite{davies2019benchmarks}, we encourage future neuromorphic hardware users to report dynamic and static energy separately to allow for fairer comparisons of algorithms. 

One of the limitations of our architecture is the need to encode zeros. Since we compute using intervals and counterweights, a zero has to be represented by two spikes from input and \emph{readout} neurons arriving at the same time so that their weights cancel each other out. A rate-coded architecture can omit sending spikes when the value is zero. The sparsity of frame-based datasets and therefore the opportunity to exploit this circumstance is highly variable, with MNIST containing 80.7\% zeros, whereas CIFAR10 only has 0.25\% zeros. In either case, even for a dataset such as MNIST, we still use orders of magnitudes fewer spikes than rate-coded techniques. 
Furthermore, relying on the precise timing of single spikes makes networks that were converted using Quartz more sensitive to noise when compared to rate coding. In the presence of transient noise sources that cause additional spikes in the network, performance is likely to degrade. In mixed-signal hardware, the degree to which hardware variability affects network performance depends on the number of layers (errors accumulate), ratio of thresholds to weights (higher is more robust) and the amount of mismatch. 

Future improvements might include a mix of rate- and temporal coding using a Time Difference Encoder~\cite{d2020event} that translates a spike time interval into a spike rate, to try to combine the best of both worlds.
A further consequence of the decoupled layers is that we can choose different time constants for each layer. Since quantization errors across the network accumulate, one could choose a larger $T_\text{max}$ in the first layer and gradually reduce it in deeper layers. In addition, porting Quartz to Loihi 2, which supports shared weights for convolutional layers on a smaller technology node, is expected to further increase efficiency.


\begin{algorithm}[ht]
\DontPrintSemicolon
\KwIn{ANN $\mathcal{A}$.\\Sample input $x$.\\Percentile $p$.}
 $u$ = \{\}\\
 \ForEach{parameter layer $l$ in $\mathcal{A}$}{
  Let $\mathcal{B}$ be a subset of $\mathcal{A}$ up until layer $l$.\\ 
  Each layer $l$ has weights $w_l$ and biases $b_l$.\\
  $\hat y = \mathcal{B}(x)$\\
  $s = p^{th} \text{ percentile of }\hat y$\\
  $w_l \leftarrow w_l / s$\\
  $u \leftarrow u \cup s$ \\
  $b_l \leftarrow b_l / \prod_i u_i$
  
}
 \KwResult{Activations of all layers in $\mathcal{A}$ are now normalized to a percentile $p$.}
 \caption{ANN activation normalization.}
 \label{alg:output-normalization}
\end{algorithm}

\section{Methods}
Starting from a sequence of layers that has been trained on frames, we follow two steps to end up with a Quartz network. The first is to normalize all activations in the network to a maximum of $\theta$, the spike firing threshold, which is detailed in the next section. The second step is then to replace all ReLU neuron with spiking equivalents which implement Equations \ref{eq:syn-current} - \ref{eq:rectifying-jump-condition}. Frame inputs have to be latency-encoded using Equation~\ref{eq:encoding}. 

\subsection{Bias-corrected Activation Normalization}
\begin{figure}[p]
    \qquad\qquad\qquad\qquad\qquad\textbf{Weight normalization\cite{rueckauer2017conversion} \qquad\qquad\qquad\qquad\qquad\qquad Bias-corrected normalization}\par\medskip
    \centering
    \includegraphics[width=0.45\columnwidth]{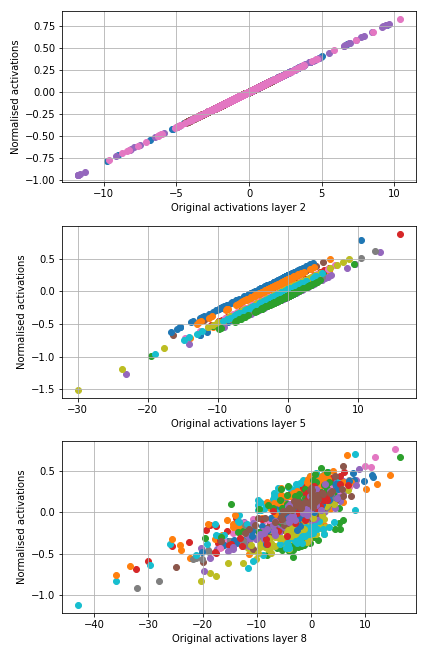}
    \vrule
    \includegraphics[width=0.45\columnwidth]{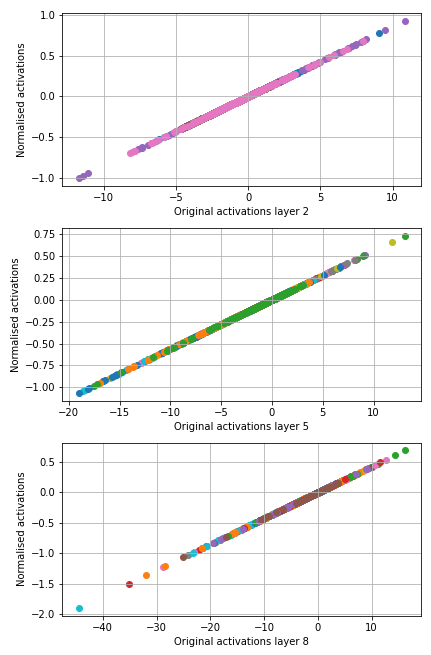}
    
    \caption{ANN normalization techniques by Rueckauer et al.~\cite{rueckauer2017conversion} (left) and us (right). Shown are 3 parameter layers (1 row each) of a VGG11 network that was trained on ImageNet. The x axis shows activations in the original ANN layer, whereas y axis shows activation for the same neurons in an activation-normalized layer. If all the activations are scaled correctly, then they should form a diagonal, which is the case for our method. Rueckauer et al.'s method does not scale biases correctly, which leads to information loss for deeper layers. Each color represents a channel, making visible the effect that the biases have.}
    \label{fig:output-normalization}
\end{figure}

Many conversion frameworks normalize ANN activations in intermediate and output layers to a maximum desired value, because the SNN activation is similarly bound to a maximum value by number of time steps per layer~\cite{diehl2015fast, sengupta2019going}. Those works make use of a popular normalization scheme proposed by Rueckauer et al.~\cite{rueckauer2017conversion}, which is named \emph{data-based weight normalization}. Using that method, ANN weights and biases are updated such that $W^l \rightarrow W^l \frac{\lambda^{l-1}}{\lambda^l}$ and $b^l \rightarrow b^l/\lambda^l$, where $\lambda^l$ is a high-percentile activation in a layer $l$. The method does not scale the biases correctly, which decreases performance especially for deeper normalized ANNs. We improve this method by correcting the term that scales the biases, which is described in Algorithm~\ref{alg:output-normalization}. Using our \emph{activation normalization} method it's possible to scale arbitrarily deep feed-forward architectures without any performance loss (up to numerical accuracy). We provide visual evidence for the efficacy of our method in Figure~\ref{fig:output-normalization}.

\begin{figure}[p]
    \centering
    \includegraphics[width=0.8\linewidth]{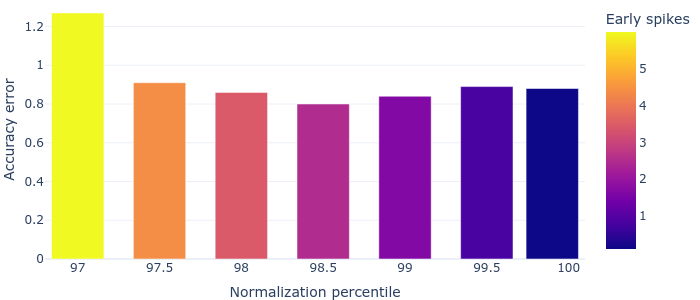}
    \caption{Classification accuracy and percentage of prematurely firing neurons as a function of normalization percentage.}
    \label{fig:early_spikes}
\end{figure}

\subsection{Effect of Normalization Percentage on Accuracy}
In rate-coded networks, the limitation of exceeding a maximum activation in the ANN can be somewhat softened by using a soft reset mechanism, therefore allowing the membrane potential to store more information across time. In temporal coding frameworks, such a soft reset is of no benefit as any neuron only fires once. Since everything depends on the single emitted spikes, such networks can be more sensitive to early firing, for example when a few large positive inputs arrive before many smaller negative ones. In an ANN, all the inputs are summed up at once and would potentially cancel each other out, but in the converted SNN the integration over time could result in a premature spike which is the major source of discrepancy between ANN and SNN performance when using Quartz. Figure~\ref{fig:early_spikes} shows the classification accuracy on MNIST for different normalization percentages, together with a color-coded percentage of early spikes in the network. As the activation in the ANN is normalized to smaller and smaller values than the maximum (100\%), the amount of early spikes increases as there occur more and more activations above threshold levels. Another source of error however is the quantization error when a high normalization value is chosen, which the data-based scheme tackles. We therefore choose a sweet spot which trades off quantization error vs percentages of early spikes in the network, for the case of MNIST a normalization percentile of $98.5$\%.





\section{Author contributions}
GL and SS developed the methods, GL and GO conducted the experiments, all authors contributed to writing the manuscript. 

\section{Data availability statement}
Our code and the models used are publicly available in two repositories: \url{https://github.com/biphasic/Quartz} and \url{https://github.com/biphasic/Quartz-on-Loihi/}

\section{Additional information}
The authors declare no competing interests.
 

\end{document}